\pdfoutput=1
\documentclass[11pt]{article}

\usepackage[final]{EMNLP2023}
\usepackage{amsmath}
\usepackage{amssymb}
\usepackage{times}
\usepackage{latexsym}
\usepackage{float}
\usepackage{booktabs}
\usepackage[T1]{fontenc}
\usepackage{graphicx}
\usepackage[utf8]{inputenc}

\usepackage{microtype}

\usepackage{inconsolata}

\title
{Cognitively-Inspired Emergent Communication via Knowledge Graphs for Assisting the Visually Impaired}
\author{ 
  \textbf{Ruxiao Chen}\textsuperscript{1}, 
  \textbf{Dezheng Han}\textsuperscript{2}, 
  \textbf{Wenjie Han}\textsuperscript{2}, 
  \textbf{Shuaishuai Guo}\textsuperscript{2}\\
  \textsuperscript{1}Johns Hopkins University \textsuperscript{2}Shandong University\\
  rchen117@jh.edu, shuaishuai\_guo@sdu.edu.cn
}

\begin{document}
\maketitle
\begin{abstract}
Assistive systems for visually impaired individuals must deliver rapid, interpretable, and adaptive feedback to facilitate real-time navigation. Current approaches face a trade-off between latency and semantic richness: natural language-based systems provide detailed guidance but are too slow for dynamic scenarios, while emergent communication frameworks offer low-latency symbolic languages but lack semantic depth, limiting their utility in tactile modalities like vibration. To address these limitations, we introduce a novel framework, Cognitively-Inspired Emergent Communication via Knowledge Graphs (VAG-EC), which emulates human visual perception and cognitive mapping. Our method constructs knowledge graphs to represent objects and their relationships, incorporating attention mechanisms to prioritize task-relevant entities, thereby mirroring human selective attention. This structured approach enables the emergence of compact, interpretable, and context-sensitive symbolic languages. Extensive experiments across varying vocabulary sizes and message lengths demonstrate that VAG-EC outperforms traditional emergent communication methods in Topographic Similarity (TopSim) and Context Independence (CI). These findings underscore the potential of cognitively grounded emergent communication as a fast, adaptive, and human-aligned solution for real-time assistive technologies. Code is available at \url{https://github.com/RuxiaoChen/VAG-EC/tree/main}.
\end{abstract}

\section{Introduction} 
{
{A}ccording to the World Health Organization, as of 2021, approximately 220 million people worldwide are blind or visually impaired, accounting for about $3\%$ of the global population~\cite{who2023blindness}. This significantly affects their ability to perform basic daily activities such as navigation, eating, and personal care, often resulting in increased dependence on caregivers and a loss of privacy and autonomy. In response, a wide range of artificial intelligence (AI)-powered wearable assistive systems have been developed to enhance the independence of visually impaired individuals. These systems generally fall into two categories: natural language-based communication and discrete learned symbolic signaling frameworks.

Natural language-based designs leverage advances in computer vision, natural language processing (NLP), and reinforcement learning to enable AI agents to provide spoken guidance grounded in visual context. This approach has been widely explored within the domain of vision-language navigation (VLN), where agents generate natural-language instructions to support user navigation through complex environments~\cite{Anderson2018cvpr, Kurita2020, Weiss2019}. While natural language offers high expressiveness and a low barrier to user understanding, it suffers from high latency, particularly in fast-changing scenarios. For example, in a situation where a bicycle rapidly approaches a visually impaired user, verbal communication may take several seconds, rendering it ineffective for timely hazard avoidance. This limitation has motivated exploration into alternative paradigms capable of enabling faster and more compact communication.

One promising alternative is \emph{Emergent Communication} (EC), wherein artificial agents develop discrete symbolic protocols through interactive learning. Compared to natural language, EC messages are compact, low-latency, and well-suited for real-time assistive interactions. However, most existing EC approaches operate directly on unstructured visual inputs or high-dimensional embeddings, failing to leverage structured, cognitively informed representations~\cite{Mu2021}. This limits their semantic alignment with how humans process visual scenes—by segmenting them into entities and reasoning over inter-object relationships~\cite{biederman1987, teney2017}. Moreover, traditional EC frameworks often treat all input elements uniformly, lacking attention mechanisms to prioritize task-relevant information. In contrast, human perception relies heavily on \emph{selective attention}—the dynamic allocation of focus to salient objects based on contextual relevance. Without such mechanisms, EC agents may produce ambiguous or overly verbose messages, which is especially problematic for constrained output channels such as vibration or haptic feedback\cite{Zhou_Hao_2024,conklin2023compositionality}.

Human spatial understanding is further grounded in the construction of \emph{cognitive maps}—internal representations that organize the environment into meaningful entities and their relationships, enabling efficient reasoning and navigation~\cite{Epstein2017CognitiveMap, Ishikaw2021}. For individuals without access to visual input, constructing such maps depends on external systems that can extract and prioritize key environmental features to guide decision-making.

To address these challenges, we propose the \textbf{VAG-EC} (\textit{Visual Attention Graph-based Emergent Communication}) framework, a cognitively inspired communication paradigm that combines structured semantic abstraction with attention-driven information filtering. VAG-EC first transforms a visual scene into a \emph{knowledge graph} that encodes object-level semantics along with spatial and functional relationships. This graph-based abstraction introduces human-aligned compositionality, enabling agents to reason over high-level semantic structures. To emulate human selective attention, VAG-EC integrates an attention mechanism that scores nodes within the knowledge graph based on task relevance. This mechanism enables the agent to focus on salient entities and encode their significance into symbolic messages via EC protocols.

By fusing cognitive structure with attention-guided selection, VAG-EC facilitates the emergence of interpretable, efficient, and context-sensitive communication protocols. These protocols are trained using a referential game with graph-structured inputs and evaluated using standard EC metrics such as \emph{Topographic Similarity} (TopSim) and \emph{Context Independence} (CI), across varying vocabulary sizes and message lengths.

Our contributions can be summarized as follows:
\begin{itemize}
    \item  We introduce a compact and interpretable alternative to natural language that supports real-time interaction in assistive scenarios.
    \item  We propose a structured encoding of visual scenes using knowledge graphs to capture human-aligned semantics and object relations.
    \item  We incorporate task-driven attention mechanisms to prioritize salient graph nodes, ensuring that generated messages are concise, relevant, and semantically meaningful.
\end{itemize}
}

\begin{figure*}
  \centering
  \includegraphics[width=16cm]{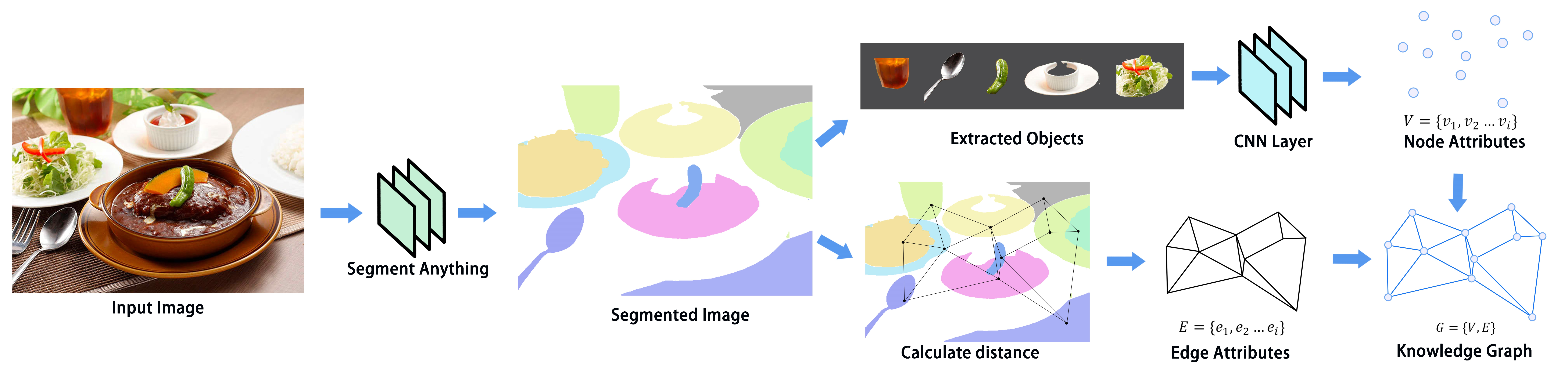}
  \caption{Pipeline for constructing a knowledge graph from a dining image. The input image is segmented using Segment Anything, followed by object extraction and feature encoding. Node attributes are derived from object embeddings, while edge attributes are computed based on spatial proximity, forming a structured graph representation of the scene.}

  \label{fig1}
\end{figure*}

% Drawing all the insights above, this article proposed an emergent semantic communication framework for MAR. The main contribution of this article can be summarized as follows:
% \begin{itemize}
% 	\item Establishing a basic idea and structure for implementing emergent communication in the field of MAR for visual data processing.
%         \item Demonstrating the emergent semantic communication's superior generalization ability and significantly smaller communication data sizes, identifying its potential benefits in industrial, societal, and business applications.
% 	\item Considering the channel uncertainty in real-world scenarios in the training process of emergent communication, we achieve increased robustness when confronted with varying channels.
% \end{itemize}

\section{Related Work}

\subsection{Emergent Communication with Structured Semantics}

EC investigates how artificial agents can autonomously develop discrete symbolic languages through interactive learning, often within referential or signaling game settings~\cite{Lewis1986}. In these paradigms, a speaker observes a target stimulus and encodes it into a symbolic message, which a listener uses to infer the intended referent from among distractors. Communication protocols emerge through reinforcement or supervised learning, optimizing task performance via message exchange. While early EC studies operated on abstract symbolic inputs such as one-hot identifiers or handcrafted attribute vectors~\cite{NEURIPS2022_093b08a7}, subsequent work extended EC to richer modalities, incorporating raw images~\cite{Dess2021}, pixel-based features~\cite{nikolaus2024emergent}, and multimodal embeddings~\cite{lee-2024-one}.

Recent efforts have sought to improve the compositionality, interpretability, and generalization of EC protocols beyond mere architectural refinements. For example, Chaabouni et al.~\cite{Chaabouni2022} demonstrated that constraining the symbol space via \emph{vocabulary bottlenecks} fosters the emergence of more compositional and transferable languages in multi-agent populations. Xu et al.~\cite{xu2022compositional} employed disentangled latent representations using $\beta$-VAE encoders to promote factorized semantic abstractions, enhancing zero-shot communication capabilities. In another direction, Nikolaus et al.~\cite{nikolaus2024emergent} introduced mechanisms for \emph{repair and recovery} in communication, whereby agents iteratively refine degraded messages through conversational feedback, leading to greater resilience under noise.

Despite these advances, most existing EC models still operate on dense, unstructured feature vectors derived from convolutional neural networks, lacking any explicit encoding of entity-level semantics or inter-object relationships. Such representations fall short in capturing the relational and compositional nature of real-world scenes, leading to messages that often reflect superficial correlations rather than meaningful semantic distinctions. Moreover, the absence of attention mechanisms results in uniform treatment of all input elements, limiting the ability to prioritize task-relevant content—a key facet of human communication, especially under real-time constraints.

\subsection{Graph-Based Visual Abstraction and Cognitive Grounding}

To bridge the gap menioned above, the vision community has increasingly adopted graph-based representations to model structured relationships within visual inputs~\cite{han2022vision, Munir_2024_CVPR, 9156945, Munir_2023_CVPR}. These models treat image patches or objects as graph nodes and represent spatial, functional, or contextual relations as edges, enabling the application of graph neural networks (GNNs) to process scene-level semantics. Such approaches inherently support non-local reasoning, compositional abstraction, and relational inference—properties that are critical for symbolic communication and decision-making.

For instance, ViG~\cite{han2022vision} models global interactions through fully connected image graphs, allowing node-level features to aggregate spatial dependencies via GNN layers. GreedyViG~\cite{Munir_2024_CVPR} builds on this by proposing an efficient graph construction mechanism using greedy axial connectivity, substantially reducing computational overhead while retaining critical structural information. These methods have demonstrated strong performance across standard benchmarks and point to the scalability and flexibility of graph-based encodings for high-resolution vision tasks.

From a cognitive standpoint, such structured visual abstractions resonate with the concept of \emph{cognitive maps}—internalized mental representations that humans use to organize spatial environments and reason about object relations~\cite{Guelton2023}. While most existing works have employed graph-based models primarily for visual recognition or segmentation~\cite{NEURIPS2023_e3cdc587, vanbergen2025}, their alignment with cognitive structures makes them particularly suitable for supporting symbolic reasoning and language emergence \cite{conklin2023compositionality}.

In this work, we extend this line of inquiry by integrating knowledge graphs as structured inputs for EC, enabling agents to reason over entities and their relationships in a cognitively aligned manner. Furthermore, we incorporate attention-driven filtering mechanisms to simulate human selective focus, allowing the emergent messages to be both concise and semantically relevant. This fusion of cognitive grounding and attention prioritization addresses the limitations of previous EC systems, paving the way for robust, real-time symbolic communication in assistive settings.

\section{Construction of Visual Cognitive Maps}

Human spatial cognition fundamentally relies on the formation of \emph{cognitive maps}—internalized mental models that encapsulate the layout and relationships among salient entities in the environment~\cite{Ishikaw2021, BEHRENS2018490}. These maps are incrementally constructed and refined through multimodal sensory input and interaction, allowing individuals to navigate, plan actions, and make inferences about spatial structure. For instance, when seated at a dining table, a person rapidly constructs a mental representation of the location of plates, cutlery, and food items, updating it dynamically through visual or haptic feedback. For individuals with visual impairments, however, constructing such maps is substantially more difficult, given the limited and often sequential nature of tactile information acquisition.

In the context of assistive AI systems, the graph-theoretic abstraction of such cognitive maps provides a principled and computationally tractable representation of structured visual environments. Inspired by human cognition, we model visual scenes as \emph{knowledge graphs}, where nodes correspond to object instances and edges encode spatial or functional relationships. This structured representation serves as the basis for downstream reasoning and communication tasks within our emergent communication framework.

To extract object-level representations from raw visual input, we employ the Segment Anything Model (SAM)~\cite{kirillov2023segany}, a general-purpose, zero-shot segmentation model capable of identifying object boundaries based on low-level visual cues such as texture and edges. Despite its strong generalization ability, SAM lacks semantic grounding and often yields redundant or spurious segments. To mitigate this, we introduce a filtering pipeline based on a combination of heuristics—including object size, spatial prominence, and contextual relevance—to retain only meaningful and distinct instances.

From the filtered set, we select the top-$N$ objects to serve as graph nodes. Their visual features are encoded using a pretrained Convolutional Neural Network (CNN), which yields a fixed-dimensional embedding for each node. Pairwise spatial relationships are encoded by constructing edges between each node and its $n$ nearest neighbors, based on the Euclidean distance between object centroids—a choice aligned with perceptual proximity priors in human cognition.

Formally, the knowledge graph for image $I_i$ is denoted as $c_i = (V_i, E_i)$, where $V_i$ is the set of selected object nodes and $E_i$ the set of proximity-based edges. This graph is processed using a Graph Convolutional Network (GCN)~\cite{8954227}, which iteratively aggregates and transforms node features by propagating information across local neighborhoods. The update rule at layer $l+1$ is defined as:
\begin{equation}
H^{l+1} = \sigma(\hat{D}^{-\frac{1}{2}} \hat{A} \hat{D}^{-\frac{1}{2}} H^l W^l),
\end{equation}
where $\hat{A} = A + I$ represents the adjacency matrix with added self-loops, $\hat{D}$ is its degree matrix, $W^l$ is a learnable weight matrix, and $\sigma(\cdot)$ denotes a non-linear activation function.

To emulate the human ability to selectively attend to task-relevant stimuli, we integrate an attention mechanism into the graph reasoning pipeline~\cite{DBLPRiUN23}. Each node $j$ is assigned a dynamic attention weight $w_j$ that reflects its contextual salience with respect to the task objective. The global representation of the graph, $J$, is then computed as an attention-weighted aggregation of node embeddings:
\begin{equation}
J = g(c_i) = \sum_j w_j h_j,
\end{equation}
where $h_j$ denotes the feature vector of node $j$, and $g(\cdot)$ represents the complete feature extraction process encompassing the GCN and attention modules.

This cognitively inspired representation pipeline enables our system to capture human-aligned compositional structures while focusing on the most pertinent aspects of the environment. It thereby forms a robust foundation for downstream symbolic communication tasks that demand both interpretability and efficiency.

\begin{figure*}[t]
  \centering
  \includegraphics[width=16cm]{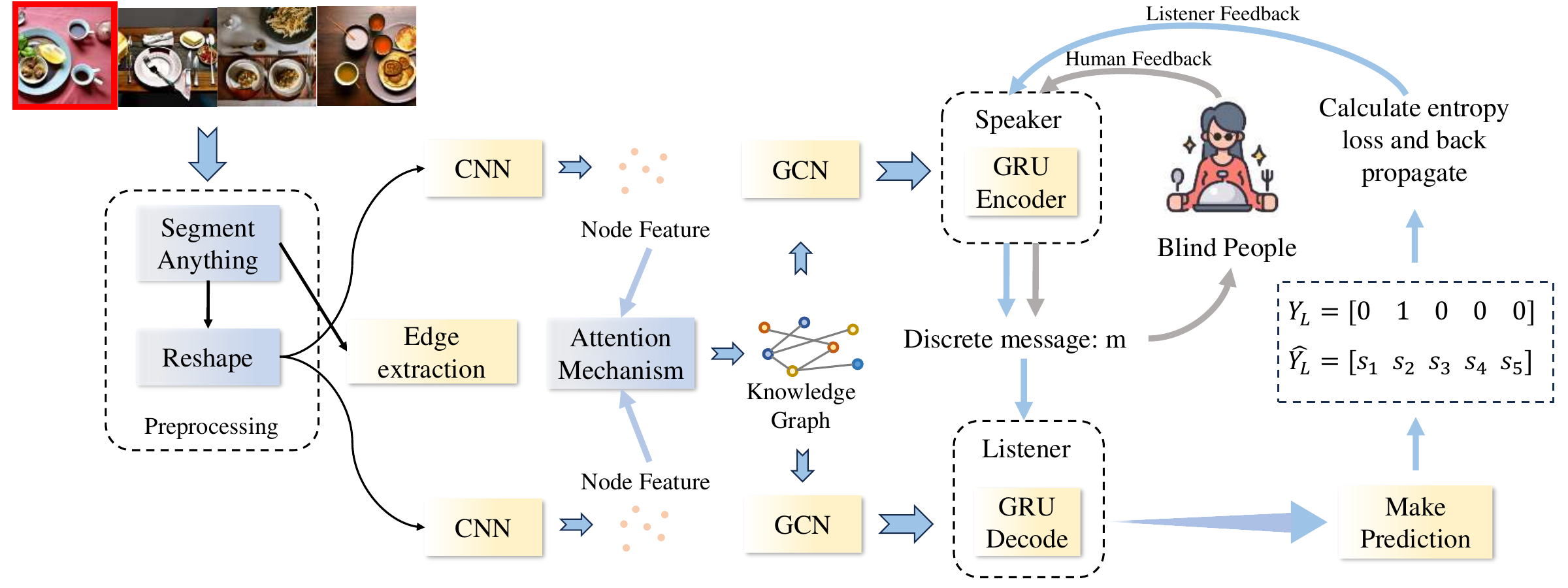}
  \caption{Overview of the proposed VAG-EC framework. Visual scenes are segmented and converted into knowledge graphs, which are encoded by the speaker to generate discrete messages. The listener decodes the message to identify the correct scene, with human feedback guiding end-to-end optimization.}
  \label{fig1}
\end{figure*}

\section{Visual‐Attention Graph-based Emergent Communication (VAG-EC)}

\subsection{Preliminary}

Our framework builds on the classical \emph{Lewis signaling game}~\cite{Lewis1986}, a two-agent communication protocol in which a \textbf{speaker} observes a world state and sends a discrete message to a \textbf{listener}, who must infer the correct state from a set of distractors.

A widely used extension of this game replaces symbolic states with natural images:  
the speaker is shown a target image, and the listener receives the same image along with distractors.  
The speaker emits a message, and the listener selects the candidate that best matches it.  
While this grounds communication in perceptual input, most implementations operate directly on raw pixels or CNN features, without any explicit structural representation.  
As a result, EC agents often develop protocols based on low-level visual cues—such as color or texture—rather than semantically meaningful concepts~\cite{han2022vision}.  
This hinders both the interpretability of the emergent language and its ability to generalize across contexts.

From a cognitive perspective, this design diverges from how humans process and reason about visual environments.  
Extensive research in psychology suggests that people construct \emph{cognitive maps}—structured, graph-like internal representations that encode objects and their interrelationships~\cite{Epstein2017CognitiveMap, Li2023}.  
These cognitive graphs serve as organizing schemata for navigating physical, social, and conceptual spaces, supporting generalization, abstraction, and flexible inference~\cite{PEER202137}.  
Moreover, they encode both observed and inferred relationships in a map-like space, enabling spatial reasoning beyond what is directly visible~\cite{Li2023}.

To align emergent communication with these cognitive principles, we construct a visual cognitive graph for each image.  
Given an input image $I_i$, we extract a knowledge graph $c_i = (V_i, E_i)$, where $V_i$ denotes object nodes (obtained via SAM segmentation) and $E_i$ encodes spatial or functional relationships between them.  
These graphs serve as structured, object-centric representations of the scene, allowing both the speaker and listener to reason over high-level semantics rather than raw visual patterns.

\subsection{Game Setup}
Specifically, the speaker and listener independently process graphs via their respective graph feature extractors, $g_{\text{s}}(\cdot)$ and $g_{\text{l}}(\cdot)$, implemented as attention-augmented GCNs.  
Given a graph $c_i$, $g_{\text{s}}(c_i)$ or $g_{\text{l}}(c_i)$ produces a feature embedding summarizing its semantic content.

During each communication round, the speaker observes the target graph $c_0$ and encodes it into $g_{\text{s}}(c_0)$.  
The listener receives a set of candidate graphs $\{c_s, c_1, \dots, c_n\}$ (one target and $n$ distractors) and processes them into embeddings $\{g_{\text{l}}(c_0), g_{\text{l}}(c_1), \dots, g_{\text{l}}(c_n)\}$.

Communication proceeds as in the classical Lewis signalling framework:  
the speaker transmits a discrete message $m$, and the listener uses it to identify the target among the candidates based on the learned embedding space \cite{ohmer2022,Guo_abs-1910-05291}.

\subsection{Speaker and Listener}

Both the speaker and listener are implemented as Gated Recurrent Units (GRUs) with separate parameters \cite{ogunleye2024emergent,Mu2021}.  
The speaker is parameterized by $\theta$ and generates a discrete message $m=(m_1,\dots,m_L)$, where each token $m_\ell\in\mathcal{V}$ and $\mathcal{V}$ is the shared vocabulary of size $|\mathcal{V}|$.  
Given the graph-structured embedding $g_{\text{s}}(c_s)$ of the target scene $c_s$, the speaker samples a message distribution:
\begin{equation}
\label{eq:speaker}
p(m\mid c_s)=f_{\text{enc}}\!\bigl(g_{\text{s}}(c_s); \theta\bigr).
\end{equation}

The listener, parameterized by $\phi$, receives the sampled message $m$ and, for each candidate scene $c_i$, computes a matching score between the decoded message and the listener's graph embedding $g_{\text{l}}(c_i)$.  
The probability that $c_i$ is the target is given by:
\begin{equation}
\label{eq:listener}
p(y_i=1\mid c_i, m) = \sigma\!\Bigl(f_{\text{dec}}(m; \phi)^\top g_{\text{l}}(c_i)\Bigr),
\end{equation}
where $\sigma(\cdot)$ denotes the logistic sigmoid function, and $y_i\in\{0,1\}$ indicates whether $c_i$ is the true target ($y_i=1$) or a distractor ($y_i=0$).

Training involves jointly optimizing the speaker and listener to maximize the likelihood of correct listener predictions.  
Let $T$ denote the set of targets, $S$ the set of sampled messages, and $G$ the set of graph embeddings for candidates.  
The loss function for a batch is:
\begin{equation}
\label{eq:loss}
\mathcal{L}(T, S, G) = - \sum_i \log p(y_i \mid c_i, \hat{m}),
\end{equation}
where $\hat{m}\sim p(m\mid c_s)$ is a message sampled from the speaker distribution conditioned on the target.

\subsection{Differentiable Message Sampling}

Since the message $m = (m_1, \dots, m_L)$ consists of discrete symbols from a vocabulary $\mathcal{V}$, the sampling process in Eq.~\eqref{eq:speaker} is non-differentiable.  
This prevents direct optimization of the speaker via gradient-based methods such as SGD.

To address this, we adopt the Gumbel–Softmax relaxation, which approximates discrete sampling using a continuous, differentiable distribution ~\cite{Mu2021, Carmeli2023}.  
Specifically, let $\pi_{\ell} \in \mathbb{R}^{|\mathcal{V}|}$ be the logits (unnormalized probabilities) output by the speaker's GRU for position $\ell$, and let $g_{\ell k} \sim \text{Gumbel}(0,1)$ be i.i.d. noise variables.  
The relaxed message $\tilde{m}_\ell \in \mathbb{R}^{|\mathcal{V}|}$ at position $\ell$ is defined component-wise as:
\begin{equation}
\tilde{m}_{\ell k} = 
\frac{\exp\big((\log \pi_{\ell k} + g_{\ell k}) / \tau \big)}
     {\sum_{j=1}^{|\mathcal{V}|} \exp\big((\log \pi_{\ell j} + g_{\ell j}) / \tau \big)},
\end{equation}
where $\tau > 0$ is the temperature parameter that controls the smoothness of the distribution.  
As $\tau \to 0$, the soft sample $\tilde{m}_\ell$ approaches a one-hot vector; as $\tau$ increases, the distribution becomes smoother.  
We set $\tau=1$ in our experiments to balance stability and approximation quality.

During training, the soft messages $\tilde{m}_\ell$ are used as input to the listener’s decoder, allowing gradients to propagate through the message generation process.  
At evaluation time, we replace $\tilde{m}_\ell$ with its one-hot argmax vector, converting the soft message back into a discrete symbol from $\mathcal{V}$.

\section{Construction of Dataset}

Due to the absence of publicly available datasets focused on dining scenarios, we construct a synthetic dataset for training and validation, while reserving real-world images exclusively for testing.  
To generate diverse and semantically rich training data, we adopt a text-to-image diffusion model~\cite{Yang2023}, which progressively refines random noise into high-quality images conditioned on textual prompts.

We design a structured prompt library centered around three key categories: \textit{food}, \textit{drink}, and \textit{tableware}.  
Prompts are created by randomly sampling and combining elements from these categories, resulting in thousands of unique scene compositions.  
Each generated image is conditioned on a randomly selected prompt, enabling a wide variety of dining scene layouts.

Importantly, diffusion models introduce inherent stochasticity—identical prompts yield visually distinct images across different sampling runs \cite{nguyen2023dataset,REUTOV2023335}.  
This property allows us to generate diverse training samples while maintaining semantic consistency, helping reduce dataset bias and improve generalization.

Further details on the prompt construction process, as well as example images from both synthetic and real-world datasets, are provided in Appendix~\ref{sec:data}.

\begin{figure*}[h]
  \centering
  \includegraphics[width=16cm]{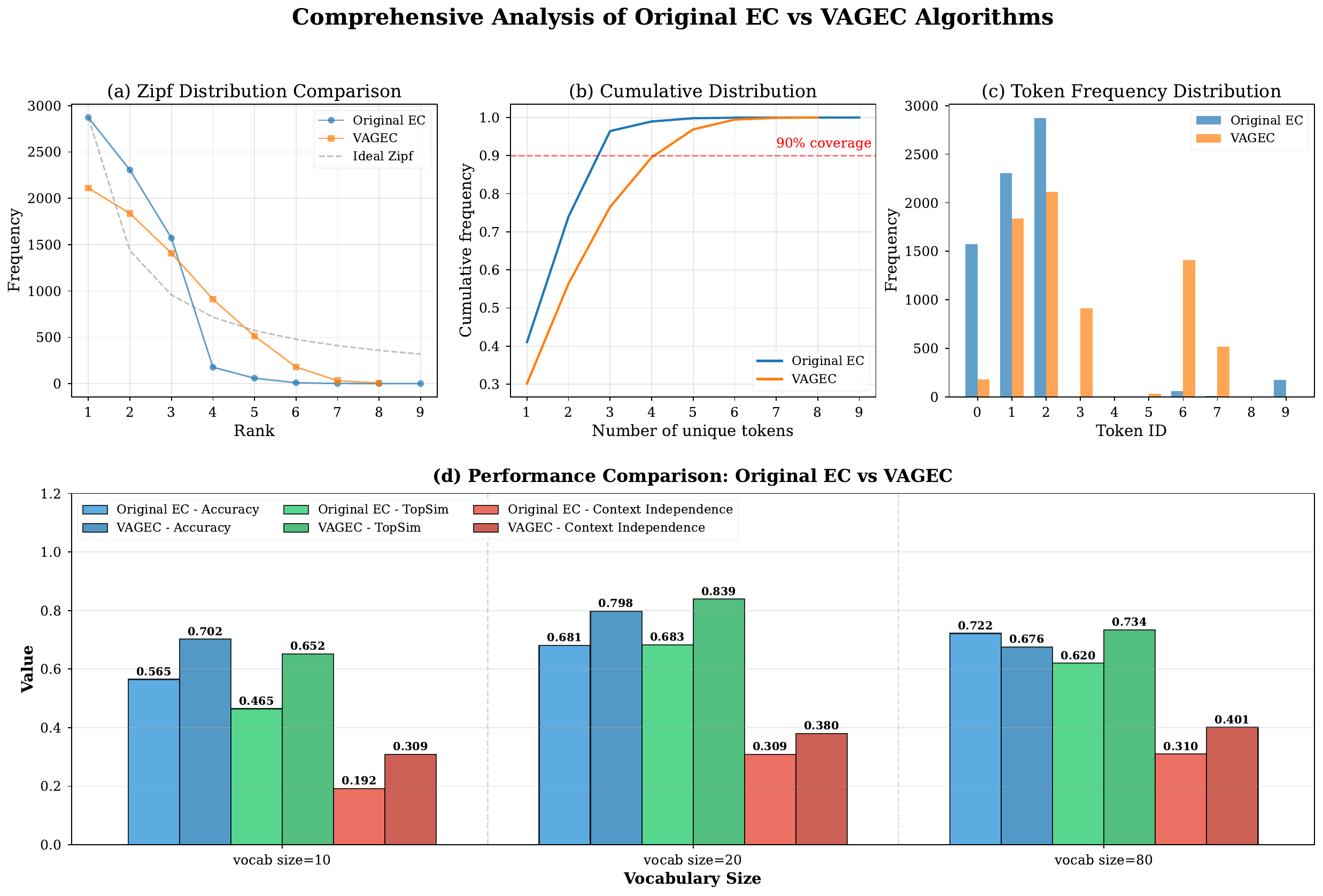}
  \caption{Quantitative comparison between the baseline EC and our proposed \textsc{VAG-EC} framework. 
(a)--(c) show token-level statistics: Zipf distribution, cumulative token coverage, and frequency histogram. 
(d) reports task-level performance across three metrics (Accuracy, TopSim, and Context Independence) under varying vocabulary sizes.}

  \label{fig4}
\end{figure*}

\section{Evaluation} 

We evaluate the effectiveness of our framework by varying the vocabulary size \(V \in \{10, 20, 80\}\) while fixing the message length \(L = 10\). This setup allows us to assess how well the model adapts to different communication bottlenecks and develops efficient symbolic protocols under constrained or relaxed settings.

To quantify the semantic quality of the learned messages, we adopt two widely used metrics in emergent communication research: Context Independence (CI) and Topographic Similarity (TopSim)~\cite{Boldt2024ARO}.  
CI measures how consistently a given token refers to the same concept across different contexts, reflecting symbolic stability and compositionality \cite{Kuci2021}.  
TopSim evaluates how well the geometric structure of the message space aligns with that of the input space, capturing the degree to which similar inputs yield similar messages. These two metrics offer complementary perspectives: CI focuses on token-level interpretability, while TopSim captures global structural alignment \cite{chaabouni2020, Peters2025}.  
Further implementation details are provided in Appendix~\ref{sec:metrics}.

To further characterize the symbolic structure of the emergent communication protocols, we analyze three distributional properties:  
(1) token frequency distribution, (2) cumulative token coverage, and (3) Zipfian behavior.
These metrics provide insight into symbolic efficiency, diversity, and naturalness.  
Token frequency distribution reveals whether the protocol suffers from \emph{token collapse} or makes full use of the vocabulary~\cite{chaabouni2019}.  
Cumulative coverage measures how many tokens are needed to account for 90\% of all message tokens, indicating expressive range~\cite{ueda-washio-2021}.  
Zipfian behavior, assessed via log--log rank-frequency plots, evaluates whether the emergent language follows the power-law regularity typical of natural languages~\cite{ueda-washio-2021, ueda2024}.

\subsection{Comparison on Accuracy, TopSim and Context Independence}

Figure\,\ref{fig4}(d) is the comparison results of \textsc{VAG-EC}  and EC baseline across all three evaluation metrics.  

For accuracy comparison, at low vocabulary sizes ($V{=}10, 20$), the baseline must reuse tokens for unrelated concepts, leading to ambiguity and degraded success.  
In contrast, \textsc{VAG-EC} builds a structured knowledge graph and focuses on \emph{goal-relevant relations} (e.g., \textit{cup, $\xleftarrow{\text{next to}}$ forks}), allowing the speaker to produce compact yet unambiguous messages.  
This semantic compression drives the accuracy gains in constrained settings.  
At $V{=}80$, the baseline improves by memorising near one-to-one image–message mappings, but this gain is superficial: TopSim and CI remain low, indicating poor generalisation.  
Since real-world assistive systems often operate under tight vocabularies, \textsc{VAG-EC}'s structural advantage is practically more valuable.

TopSim evaluates whether similar inputs yield similar messages.  
Our graph-based encoder, with attention over structured object–relation nodes, creates disentangled scene representations where compositional variations (e.g., object identity vs. color) remain linearly organised.  
This smooth latent geometry naturally improves TopSim.

CI measures whether tokens maintain consistent meanings across contexts.  
The baseline, trained on raw features, often develops polysemous symbols—especially with small $V$.  
In contrast, \textsc{VAG-EC} emits messages over pre‑factorised graph elements, enforcing semantic alignment and reducing token ambiguity.  
This explains the observed 30–60\% CI improvement across all setups.

\subsection{Message–Level Behaviour of \textsc{VAG-EC}}

Figure\,\ref{fig4} (a)-(c) presents three views of message-level token usage, comparing VAG-EC with the EC baseline. 

In the Zipf plot, natural languages typically exhibit a smooth inverse–rank curve. As we can see in Figure\,\ref{fig4}(a) , the baseline's curve drops steeply, indicating token collapse: a few symbols dominate, while the rest are underused. In contrast, VAG-EC's distribution is flatter and closer to the ideal Zipf trend. By grounding utterances in $\langle$\textit{object, relation}$\rangle$ tuples and attending over graph nodes, our model spreads usage across mid-rank tokens, producing a more compact yet expressive code.

The cumulative coverage plot quantifies lexical diversity. The baseline achieves 90\% coverage with just 4 tokens, whereas VAG-EC requires 6–7—reflecting broader symbol usage. Because each token in VAG-EC maps to a structured semantic element, the model naturally rotates through more of the vocabulary without relying on overloaded or generic symbols.

Finally, the raw frequency histogram confirms this effect at the token level: the baseline concentrates usage in IDs 0–2, while VAG-EC distributes usage more evenly and activates long-tail tokens. This balance is a direct result of grounding; with a fixed semantic schema, token meanings are stable and purpose-specific, reducing redundancy and ambiguity—key properties for downstream deployment in constrained communication channels.

\section{Conclusion}
The VAG-EC framework represents a significant advancement in assistive communication for visually impaired individuals by integrating knowledge graphs and attention mechanisms to overcome the limitations of existing technologies. Through systematic evaluation, the framework has demonstrated enhanced expressiveness and adaptability, offering a more structured and interpretable communication method. By constructing cognitive maps that mimic human visual perception, the framework facilitates efficient spatial information processing. This innovative approach not only addresses current challenges but also lays the groundwork for future research in emergent semantic communications.

\section{Limitations}

While our KG-AEC framework introduces a cognitively inspired approach to emergent communication, several limitations remain that point to future directions.

First, our current work focuses exclusively on structured dining scenes, where the object categories and spatial relations are relatively constrained. Generalizing to a broader range of scenarios—such as bathrooms (e.g., locating a toothbrush near a sink) or indoor navigation (e.g., avoiding moving obstacles)—requires handling significantly more diverse semantic content. When multiple domains are combined into a single model, scene representations may become semantically ambiguous, making it harder for agents to extract consistent and relevant relational cues \cite{rita2022emergent,Feng_An_Lu_2024}.

Second, in our current referential game setup, both training and evaluation are performed using simulated agents. In practice, the most faithful way to optimize and assess an assistive communication system is through real blind participants in a human-in-the-loop paradigm. Such an approach would allow the emergent protocol to co-evolve with user preferences, tactile interpretation, and cognitive expectations. However, involving human subjects raises ethical, logistical, and methodological complexities beyond the scope of this paper \cite{Holzinger2022HITL}. Moreover, designing and evaluating such a closed-loop human–agent system is substantial enough to merit a dedicated study. This work provides theoretical and experimental groundwork toward that broader goal.

\bibliography{anthology,custom}
\bibliographystyle{acl_natbib}

% \appendix

% \section{Example Appendix}
% \label{sec:appendix}

% This is a section in the appendix.

\clearpage

\appendix

\section{Dataset Generation}
\label{sec:data}

To simulate diverse and semantically structured dining environments, we construct a synthetic dataset using a compositional prompt template tailored for diffusion-based image generation. Each prompt specifies the spatial arrangement and object types in the scene, using the following template:

\texttt{``The center of the picture is a plate with \{food\} on this plate, a glass of \{drink\} on the \{direction1\} of the plate, a pair of \{tool1\} and \{tool2\} on the \{direction2\} of the plate, and a \{tool3\} next to the \{drink\}. The picture is photographic.''}

Here, \{\texttt{food}\}, \{\texttt{drink}\}, and \{\texttt{tool}\} are sampled from predefined category-specific vocabularies to ensure semantic consistency, while \{\texttt{direction1}\} and \{\texttt{direction2}\} are drawn from \textit{left}, \textit{right}, \textit{top}, and \textit{bottom}, introducing controlled spatial variability.

\begin{figure}[h]
  \centering
  \includegraphics[width=1\linewidth]{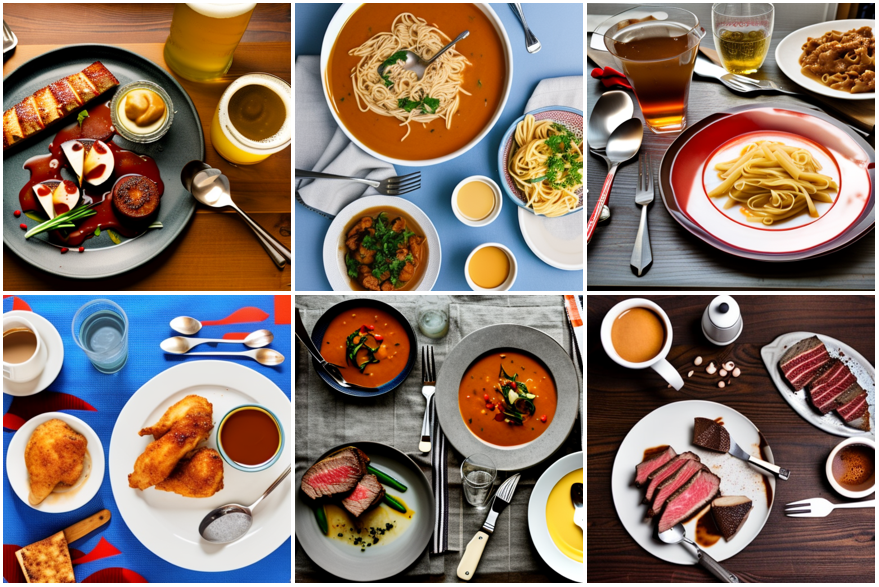}
  \caption{Example of a generated dining scenario from the synthetic dataset.}
  \label{figgf}
\end{figure}

\begin{figure}[h]
  \centering
  \includegraphics[width=1\linewidth]{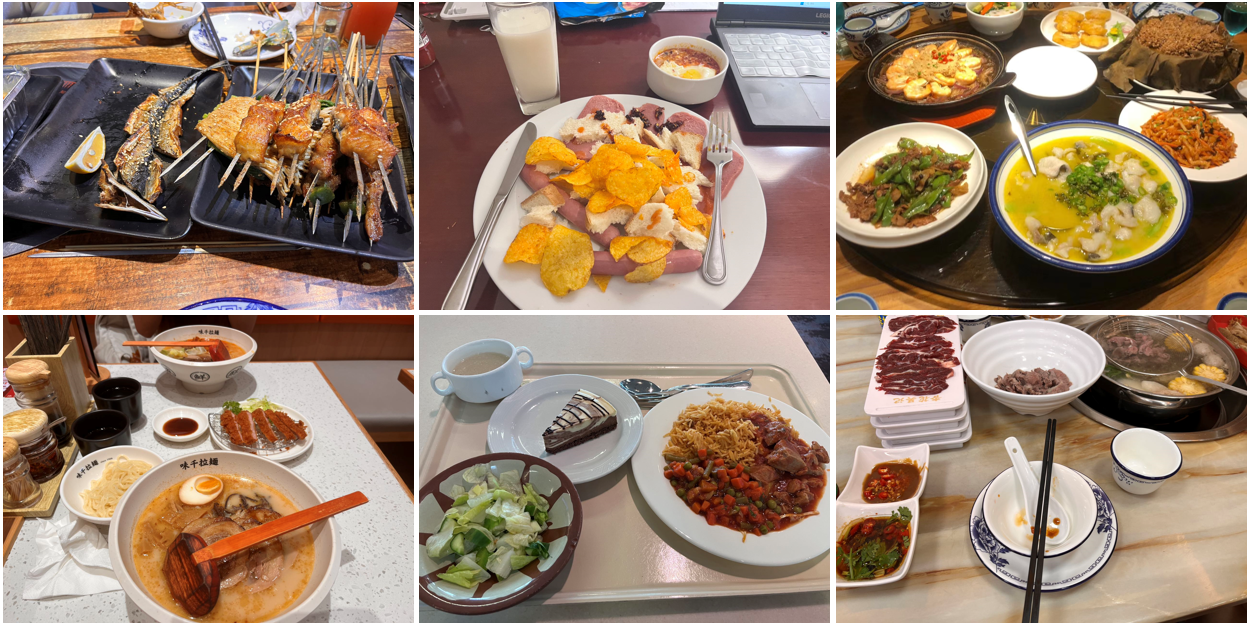}
  \caption{Example of a real-world dining scenario used for testing.}
  \label{figrf}
\end{figure}

Figure~\ref{figgf} shows an example of a generated dining image. For evaluation, we rely on real-world dining scenes (see Figure~\ref{figrf}) to test generalization beyond the synthetic domain.

\vspace{-0.5em}

\clearpage

\section{Evaluation Metrics Explanation}
\label{sec:metrics}
\subsection{Context Independence}
Context Independence (CI) measures the extent to which the generated messages remain consistent and generalizable across different environments. This is particularly important for blind individuals, as the language must be interpretable regardless of variations in the surrounding context. Given a message $m$ and a corresponding set of concepts $C$, we define context independence as:

\begin{equation} CI(C, M) = \frac{1}{|C|} \sum_{c \in C} p_m(m^c | c) \cdot p_c(c | m^c), \end{equation}

where $m^c = \arg\max_m p_c(c | m)$ is the most likely message assigned to a given concept $c$. This formulation ensures that the generated messages are semantically stable and do not fluctuate based on minor environmental variations. A higher CI score indicates that the emergent language is more reliable and less context-dependent, allowing blind users to interpret messages consistently across different situations.

Our results show that the knowledge graph-based GESC framework consistently improves CI across various experimental configurations. This suggests that encoding semantic object relationships within a graph structure leads to more stable and interpretable messages, reducing the influence of dataset-specific contextual variations.

\subsection{Topographic Similarity}
TopSim measures the structural alignment between the object space and the message space, ensuring that semantically similar objects are represented with similar messages\cite{lazaridou2018emergence}. We compute TopSim by comparing the pairwise distances in both spaces using Spearman's rank correlation coefficient:

\begin{equation} \text{TopSim} = \rho(D_c, D_m), \end{equation}

where $D_c$ represents the pairwise distances between concept embeddings and $D_m$ represents the pairwise distances between message embeddings. The Spearman correlation coefficient $\rho$ quantifies the monotonic relationship between these two distance matrices. A higher TopSim score indicates that the emergent language effectively preserves the structural relationships among objects, allowing blind individuals to infer spatial organization and object interactions more accurately.

Our experimental results demonstrate that the knowledge graph-based GESC framework achieves higher TopSim scores compared to the baseline emergent communication model. This improvement suggests that leveraging knowledge graphs enables the system to better capture and retain object relationships, leading to more structured and meaningful message representations.

\end{document}